\documentclass[conference]{IEEEtran}
\IEEEoverridecommandlockouts
\usepackage{cite}
\usepackage{amsmath,amssymb,amsfonts}
\usepackage{algorithmic}
\usepackage{graphicx}
\usepackage{textcomp}
\usepackage{xcolor}
\usepackage{multirow}
\usepackage{array}
\usepackage{subfigure}
\usepackage{forest}
\usepackage{tikz}
\usepackage{tabulary}

\usepackage{balance}

\usetikzlibrary{shapes, arrows.meta, positioning}
\def\BibTeX{{\rm B\kern-.05em{\sc i\kern-.025em b}\kern-.08em
    T\kern-.1667em\lower.7ex\hbox{E}\kern-.125emX}}
\begin{document}

\title{From Black Box to Insight: Explainable AI \\ for Extreme Event Preparedness}

\author{\IEEEauthorblockN{Kiana Vu\IEEEauthorrefmark{1}}
\IEEEauthorblockA{\textit{Department of Cybersecurity} \\
\textit{University at Albany, SUNY}\\
Albany, NY, USA \\
kvu@albany.edu}
\and
\IEEEauthorblockN{İsmet Selçuk {\"O}zer}
\IEEEauthorblockA{\textit{Department of Cybersecurity} \\
\textit{University at Albany, SUNY}\\
Albany, NY, USA \\
iozer@albany.edu}
\and
\IEEEauthorblockN{Phung Lai}
\IEEEauthorblockA{\textit{Department of Cybersecurity} \\
\textit{University at Albany, SUNY}\\
Albany, NY, USA \\
lai@albany.edu}
\and
\IEEEauthorblockN{Zheng Wu}
\IEEEauthorblockA{\textit{Dept. of Atmospheric \& Environmental Sciences} \\
\textit{University at Albany, SUNY}\\
Albany, NY, USA \\
zwu26@albany.edu}
\and
\IEEEauthorblockN{Thilanka Munasinghe}
\IEEEauthorblockA{\textit{Lally School of Management} \\
\textit{Rensselaer Polytechnic Institute}\\
Albany, NY, USA \\
thilankawillbe@gmail.com}
\and
\IEEEauthorblockN{Jennifer Wei}
\IEEEauthorblockA{\textit{Goddard Space Flight Center} \\
\textit{NASA}\\
Greenbelt, MD, USA \\
jennifer.c.wei@nasa.gov}

\thanks{\IEEEauthorrefmark{1} Corresponding author}
}
\newcommand\kv[1]{\textcolor{red}{ #1}}

% TikZ box styles
\tikzstyle{my-box}=[rectangle, rounded corners, align=left, text opacity=1, minimum height=0.5em, minimum width=1em, inner sep=2pt, fill opacity=.8, line width=0.8pt]
\tikzstyle{leaf-head}=[my-box, draw=gray!80, fill=gray!15, text=black, font=\tiny, inner xsep=2pt, inner ysep=4pt]

\tikzstyle{green-box-lv2}=[my-box, draw=green!70, fill=green!15, text=black, font=\tiny, inner xsep=2pt, inner ysep=4pt]
\tikzstyle{green-box-lv3}=[my-box, draw=green!70, fill=gray!15, text=black, font=\tiny, inner xsep=2pt, inner ysep=3.5pt]
\tikzstyle{green-box-lv4}=[my-box, draw=green!100, fill=white, text=black, font=\tiny, inner xsep=3pt, inner ysep=3.5pt]

\tikzstyle{orange-box-lv2}=[my-box, draw=orange!70, fill=orange!15, text=black, font=\tiny, inner xsep=2pt, inner ysep=4pt]
\tikzstyle{orange-box-lv3}=[my-box, draw=orange!70, fill=gray!15, text=black, font=\tiny, inner xsep=2pt, inner ysep=3.5pt]
\tikzstyle{orange-box-lv4}=[my-box, draw=orange!100, fill=white, text=black, font=\tiny, inner xsep=3pt, inner ysep=3.5pt]

\maketitle

\begin{abstract}
As climate change accelerates the frequency and severity of extreme events such as wildfires, the need for accurate, explainable, and actionable forecasting becomes increasingly urgent.  While artificial intelligence (AI) models have shown promise in predicting such events, their adoption in real-world decision-making remains limited due to their black-box nature, which limits trust, explainability, and operational readiness. 
This paper investigates the role of explainable AI (XAI) in bridging the gap between predictive accuracy and actionable insight for extreme event forecasting. Using wildfire prediction as a case study, we evaluate various AI models and employ SHapley Additive exPlanations (SHAP) to uncover key features, decision pathways, and potential biases in model behavior. Our analysis demonstrates how XAI not only clarifies model reasoning but also supports critical decision-making by domain experts and response teams. 
In addition, we provide supporting visualizations that enhance the interpretability of XAI outputs by contextualizing feature importance and temporal patterns in seasonality and geospatial characteristics. This approach enhances the usability of AI explanations for practitioners and policymakers. Our findings highlight the need for AI systems that are not only accurate but also interpretable, accessible, and trustworthy, essential for effective use in disaster preparedness, risk mitigation, and climate resilience planning.
\end{abstract}


\vspace{-2mm}
\section{Introduction}
The frequency and severity of extreme events, such as wildfires, heatwaves, and floods, have increased significantly due to climate change \cite{stott2016climate, robinson2021climate}, posing serious risks to ecosystems \cite{parmesan2000impacts}, public health \cite{ebi2021extreme}, infrastructure \cite{gough2019vulnerability}, and communities. This urgency has led to growing use of AI and machine learning (ML) for spatiotemporal forecasting, such as  wildfire and  spread prediction \cite{camps2025artificial}. 
  
Despite these advances, the real-world adoption of AI-driven forecasting tools in critical domains like disaster response remains limited.  A key barrier is the ``black-box''  nature of many AI models, like deep neural networks, which make accurate predictions but lack transparency. This poses serious challenges for decision-makers in high-stakes environments like  emergency response, firefighting, and public safety. Without clear insight into why a model makes certain predictions, it becomes difficult to build trust, assess risks, validate outcomes,  and effectively use model outputs in operations.

To address these challenges, our work adopts NASA's FAIRUST principles \cite{diventi2023nasa}, which emphasize that AI applications and data should be Findable, Accessible, Interoperable, Reusable, Understandable, Secure, and Trustworthy. We specifically focus on leveraging explainable AI (XAI) techniques to make complex model behavior more Understandable. XAI sheds light on how models use input features, revealing decision logic and potential biases. In the context of extreme event forecasting, such as wildfires, XAI helps bridge the gap between predictive accuracy and real-world usability, empowering domain experts to evaluate model trustworthiness, identify critical risk factors, and make well-informed decisions under conditions of uncertainty.

In this paper, we explore the use of  XAI in wildfire forecasting, which is a critical and complex challenge in extreme event prediction. \textit{Our contributions are threefold.} \textbf{(1)} We evaluate several state-of-the-art AI models and use SHapley Additive exPlanations (SHAP) \cite{lundberg2017unified} to interpret predictions, highlighting key features and enabling a deeper understanding of model behavior, strengths, and limitations.  \textbf{(2)} To enhance interpretability, we introduce visualizations that map feature importance across spatial and seasonal dimensions, making AI-generated insights more accessible to non-technical stakeholders, including emergency planners and policymakers.  \textbf{(3)} We show how these explainability tools can support actionable insights for extreme event preparedness, bridging the gap between model performance and operational utility. Our results emphasize the role of XAI  in trustworthy climate risk forecasting and responsible AI research with societal impact. 

\vspace{-2mm}
\section{Systematization of Extreme Weather Events Prediction and Explanation}
\begin{figure}[t]
      \centering %\{-10pt}
       \includegraphics[scale=0.26]{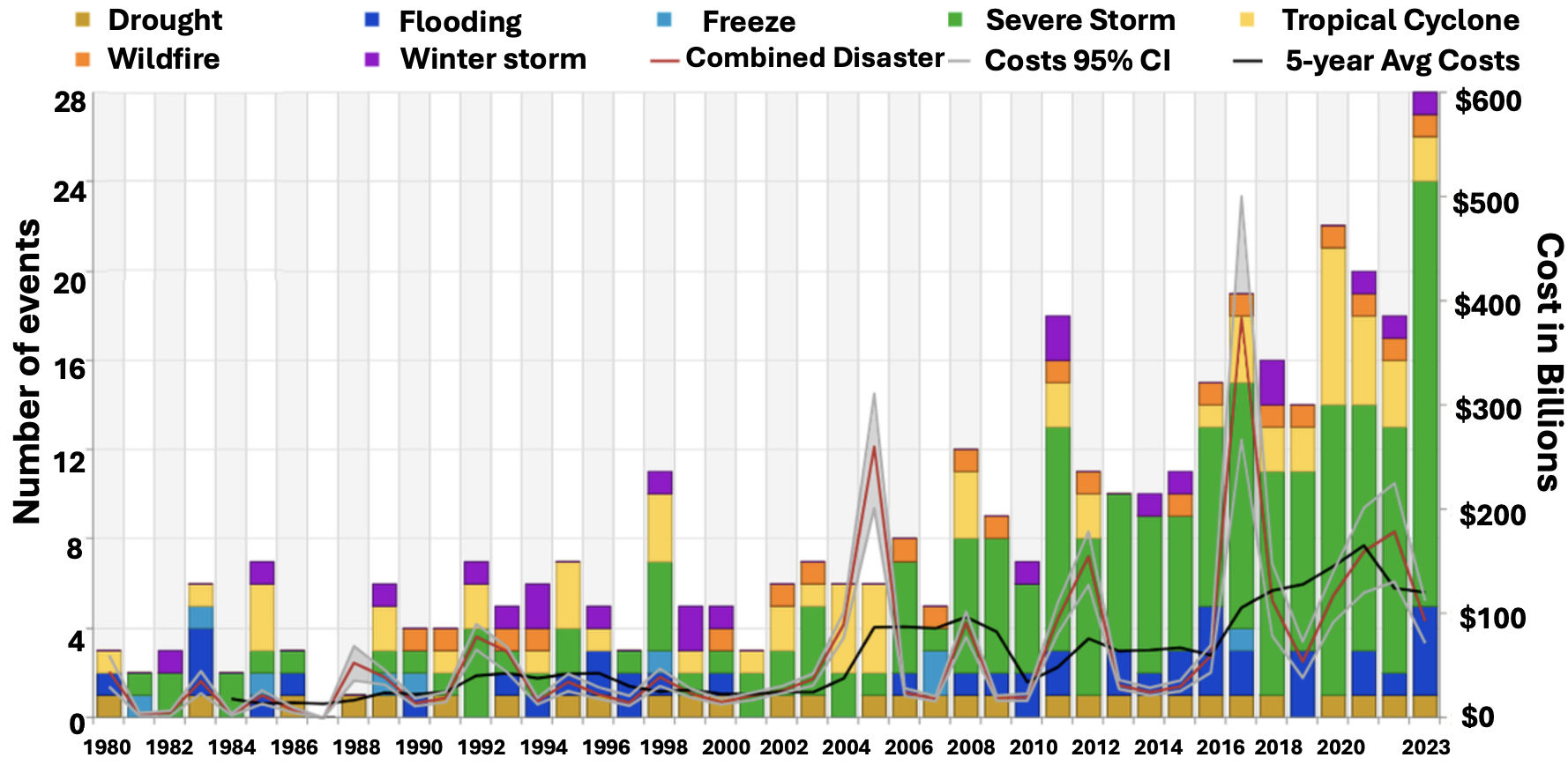} 
      \caption{Costs of U.S. extreme disaster events over time \cite{disasters}.}
      \label{fig:disaster}
\end{figure}

\subsection{Extreme Weather Events}
Extreme weather events, such as wildfires, heatwaves, hurricanes, floods, and droughts,  are statistically rare but highly impactful phenomena that deviate significantly from normal conditions \cite{ebi2021extreme}. Their impacts and associated costs have been rising, reaching hundreds of billions of dollars in recent years in the U.S. \cite{disasters} (Figure \ref{fig:disaster}).  In 2024, the National Oceanic and Atmospheric Administration (NOAA) reported 27 U.S. weather and climate disasters, each causing over one billion in damages \cite{ncei2025}. Climate change has increased the frequency and severity of these events by warming the atmosphere, which holds more moisture and intensifies heat and precipitation extremes \cite{ipcc2023synthesis}. These escalating costs and increasing frequency and intensity of extreme events have motivated significant research into understanding and predicting extreme events \cite{camps2025artificial}. Skillful forecasts are crucial for mitigation, adaptation, early warning systems, and building resilience in vulnerable communities. However, the chaotic nature of the climate system and its complex interactions across atmospheric, oceanic, and land processes pose major challenges for reliable prediction.

Wildfires are among the most destructive extreme events, becoming more frequent and intense   due to climate change and drawing increased scientific attention. These complex phenomena are influenced by weather conditions (e.g., temperature, humidity, wind)  and human activity. Wildfires can threaten lives, infrastructure, ecosystems, and economies \cite{budget2023global}. Advancing wildfire understanding and prediction is, therefore, essential and forms the focus of this paper.

\subsection{AI for Extreme Weather Events Prediction}
Despite challenges in predicting extreme weather, advances in understanding its dynamics and development of physics-based models have improved forecast skills in recent decades \cite{domeisen2022advances, lyu2023improving}. Recent AI/ML developments have further transformed forecasting by leveraging diverse data. Models such as Tree-based Ensemble, Recurrent Neural Network (RNN), Convolutional Neural Network (CNN), and Transformer have shown promise in detecting extreme weather patterns, aiding predictions of storms, floods, heatwaves, and wildfires. Ensemble and tree-based models (e.g., Random Forests, Gradient Boosting) are widely applied to floods, heatwaves, and severe storms, offering robustness to noisy data and supporting uncertainty quantification \cite{ liu2021classified,xiao2024long}. Temporal models like RNNs, Long Short-Term Memory (LSTM), and Gated Recurrent Unit (GRU), are well-suited for capturing long-term dependencies in sequential weather data, aiding in the prediction of events such as heatwaves and droughts \cite{chattopadhyay2020analog, dikshit2021improved}. More recently, Transformer-based models with attention mechanisms have shown strong performance by dynamically weighting relevant meteorological features and time steps, improving forecasts of complex events like heatwaves and hurricanes \cite{jiang2023transformer}.

% While physics-based models like WRF-Fire \cite{coen2013wrf} are widely used, they are sensitive to initial conditions and computationally demanding. AI/ML models offer a promising alternative by capturing complex, nonlinear patterns in diverse datasets to overcome the limitations of physics-based  models  \cite{camps2025artificial}. This study focuses on AI models for predicting wildfires, one of the most complex forecasting challenges. Accurate timing and location predictions are critical for effective response.  In addition to prediction, we investigate model interpretability and forecasting opportunities, which remain underexplored.

Physics-based models like WRF-Fire \cite{coen2013wrf} are widely used but computationally expensive and sensitive to initial conditions. Meanwhile, AI/ML models address these limits by capturing complex, nonlinear patterns \cite{camps2025artificial}. We focus on AI for wildfire prediction, one of the most challenging forecasting tasks, where accurate timing and location are vital. Beyond prediction, we examine interpretability and forecasting opportunities, which remain underexplored.

\subsection{XAI for Extreme Weather Events Prediction} \label{xai}

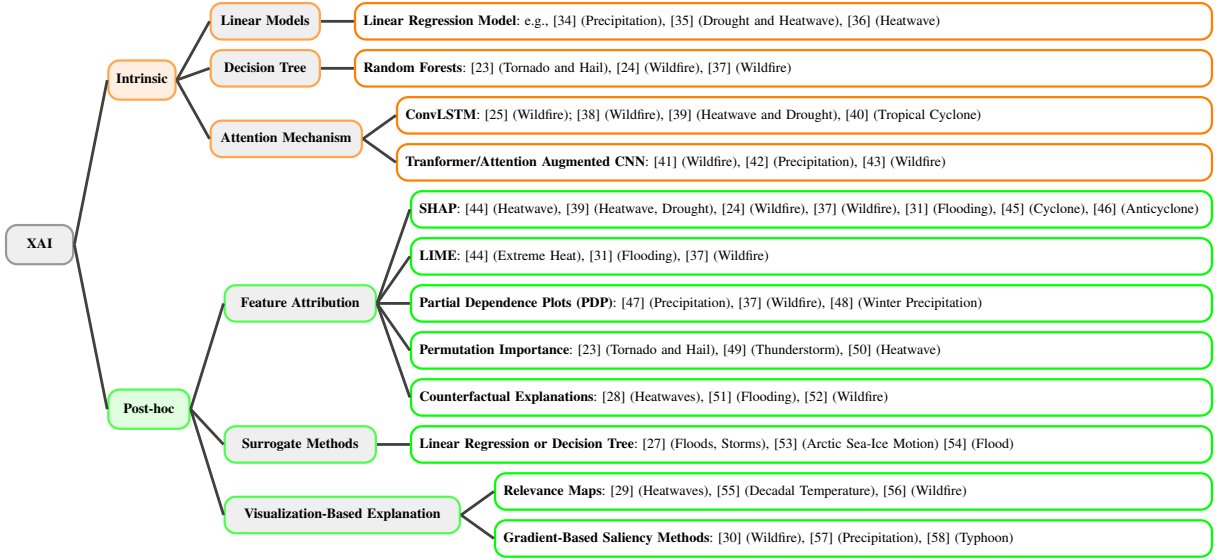
\begin{figure*}[t!]
    \centering
    \resizebox{.9 \textwidth}{!}{
        \begin{forest}
            for tree={
                grow=east,
                reversed=true,
                anchor=base west,
                parent anchor=east,
                child anchor=west,
                base=center,
                rectangle,
                rounded corners,
                align=left,
                text centered,
                minimum width=4em,
                edge+={darkgray, line width=1pt},
                s sep=3pt,
                inner xsep=2pt,
                inner ysep=3pt,
                line width=0.8pt,
                ver/.style={rotate=90, child anchor=north, parent anchor=south, anchor=center},
            },
            where level=1{text width=8em,font=\smallsize,}{},
            where level=2{text width=8em,font=\smallsize,}{},
            where level=3{text width=8em,font=\smallsize,}{},
            where level=4{text width=6em,font=\smallsize,}{},
            % Level 1
            [\centerline{\textbf{XAI}}, leaf-head, text width=2em
                % Level 2
                [\centerline{\textbf{Intrinsic}}, orange-box-lv2, text width=2em
                    [\centerline{\textbf{Linear Models}}, orange-box-lv3, text width=3.5em
                        [\text{\textbf{Linear Regression Model}: e.g., \cite{herman2018dendrology} (Precipitation), \cite{mukherjee2020compound} (Drought and Heatwave), \cite{amoatey2025evaluating} (Heatwave)}, orange-box-lv4, text width=30em]
                    ]
                    [\centerline{\textbf{Decision Tree}}, orange-box-lv3, text width=3.5em
                        [\text{\textbf{Random Forests}:
                        \cite{loken2022comparing} (Tornado and Hail), \cite{cilli2022explainable} (Wildfire), \cite{fan2024explainable} (Wildfire)}, orange-box-lv4, text width=30em]
                    ]
                    [\centerline{\textbf{Attention Mechanism}}, orange-box-lv3, text width=5em
                        [\text{\textbf{ConvLSTM}: \cite{masrur2024capturing} (Wildfire); \cite{andrianarivony2024machine} (Wildfire), \cite{pellicer2024explainable} (Heatwave and Drought), \cite{liu2016application} (Tropical Cyclone)}, orange-box-lv4, text width=28.5em]
                        [\text{\textbf{Tranformer/Attention Augmented CNN}: \cite{marjani2024application} (Wildfire), \cite{huang2022extreme} (Precipitation), \cite{prapas2023televit} (Wildfire)}, orange-box-lv4, text width=28.5em]
                    ]
                ]
                [\centerline{\textbf{Post-hoc}}, green-box-lv2, text width=2.5em
                    [\centerline{\textbf{Feature Attribution}}, green-box-lv3, text width=5em
                    [\text{\textbf{SHAP}:
                    \cite{shafiq2025extreme} (Heatwave), \cite{pellicer2024explainable} (Heatwave, Drought), \cite{cilli2022explainable} (Wildfire), \cite{fan2024explainable} (Wildfire), \cite{peng2025evaluating} (Flooding), 
                    \cite{liu2023evaluation} (Cyclone), \cite{zhang2024using} (Anticyclone)
                    %, \\ \cite{zhang2024using} (Atmospheric Blocking), \cite{reulen2024ga} (Compound Dry and Hot Events) 
                   }, green-box-lv4, text width=28em
                    ]
                    [\text{\textbf{LIME}:   \cite{shafiq2025extreme} (Extreme Heat), \cite{peng2025evaluating} (Flooding), \cite{fan2024explainable} (Wildfire)}, green-box-lv4, text width=28em
                    ]
                     [\text{\textbf{Partial Dependence Plots (PDP)}: \cite{gibson2021training} (Precipitation), 
                     \cite{fan2024explainable} (Wildfire), \cite{ibebuchi2025uncertainty} (Winter Precipitation)}, green-box-lv4, text width=28em
                    ]
                     [\text{\textbf{Permutation Importance}:  \cite{loken2022comparing} (Tornado and Hail), \cite{molina2021benchmark} (Thunderstorm), \cite{leach2021forecast} (Heatwave)}, green-box-lv4, text width=28em
                    ]
                    [\text{\textbf{Counterfactual Explanations}:  \cite{trok2024machine} (Heatwaves), \cite{chen2025counterfactual} (Flooding), \cite{thompson2023avoided} (Wildfire)}, green-box-lv4, text width=28em
                    ]
                    ][\centerline{\textbf{Surrogate Methods}}, green-box-lv3, text width=5em
                        [\text{\textbf{Linear Regression or Decision Tree}:
                        \cite{ronco2023exploring} (Floods, Storms), \cite{hoffman2025evaluating} (Arctic Sea-Ice Motion)\cite{fraehr2024assessment} (Flood)}, green-box-lv4, text width=28em
                        ]
                    ]
                    [\centerline{\textbf{Visualization-Based Explanation}}, green-box-lv3, text width=8em
                        [\text{\textbf{Relevance Maps}: \cite{wei2025xai4extremes} (Heatwaves), \cite{bommer2024finding} (Decadal Temperature), \cite{zhou2025comparative} (Wildfire)}, green-box-lv4, text width=25em
                    ]
                    [\text{\textbf{Gradient-Based Saliency Methods}: \cite{camps2024ai} (Wildfire), \cite{chen2024machine} (Precipitation), \cite{higa2021domain} (Typhoon)}, green-box-lv4, text width=25em
                    ]
                    ]
                ]
            ]
        \end{forest} }
    \caption{Explainable AI (XAI) Techniques.}
    \label{figure:Trustworthiness}
\end{figure*}

Despite their advantages, AI models often lack transparency due to their ``black-box'' nature, posing challenges for operational deployment. This underscores the need for  XAI, which helps identify key features and patterns behind predictions. In climate science, interpretability not only fosters user trust but also uncovers insights consistent with physical understanding \cite{camps2025artificial}.
XAI methods to explain AI-driven forecasts for extreme weather events generally fall into two categories (Figure~\ref{figure:Trustworthiness}): intrinsic interpretable model designs and post-hoc techniques. 

First, intrinsic XAI methods rely on models that are transparent by design \cite{rudin2019stop}. Examples include linear regression, which shows feature influence via coefficients \cite{yang2024interpretable}, and decision trees or random forests, which reveal decision paths through feature thresholds \cite{loken2022comparing, cilli2022explainable}. These models are often used in extreme event prediction due to their simplicity and interpretability. Attention-based deep learning models  can also be intrinsically interpretable. For example, \cite{masrur2024capturing} uses a Transformer with multi-head self-attention to identify key temporal and spatial drivers in next-day wildfire prediction.

Second, post-hoc XAI methods are applied after model training to explain predictions from complex AI/ML models. Common techniques include feature attribution like SHAP \cite{lundberg2017unified}, LIME \cite{ribeiro2016should},  surrogate model \cite{ronco2023exploring}, 
counterfactual explanations \cite{trok2024machine, camps2025artificial}, 
and visualization-based methods \cite{wei2025xai4extremes, camps2024ai}. SHAP is widely used in climate science for offering both global and local interpretability. In extreme weather forecasting, SHAP reveals key atmospheric features, enhancing trust and insight \cite{ peng2025evaluating}. It helps assess wildfire risk \cite{cilli2022explainable} and finds critical factors like humidity and wind \cite{abdollahi2023explainable}.

Most extreme weather prediction studies favor post-hoc XAI methods, as they explain complex AI models better than   inherently interpretable models. Since high-performing models are often opaque, post-hoc tools like SHAP provide critical insights, build trust, and support informed planning \cite{liao2025tackling}. In this work, we use SHAP to interpret wildfire  predictions,  ensuring explanations are physically meaningful and informative.

% Summary figure for XAI 

\vspace{-2mm}
\section{Background} 
\subsection{AI Prediction Models}
To predict extreme events, AI models typically use multiple weather and atmospheric features over a period preceding the event. Each feature $x_i$ is   a time series $\{x_{it}\}_{t=1}^L$, where $L$ is the length of the time window. An input sample $x$ consists of time series data of all $N$ features, denoted as $x = \{x_{i}\}_{i=1}^N$. This formulation captures the evolving dynamics of multiple environmental factors over time, which are essential for modeling complex phenomena like wildfire occurrence. 

An AI model prediction $f(x)$ takes a time series $x$ as input and produces a binary prediction $y \in \{0,1\}$, where $y=1$ indicates the predicted occurrence of a wildfire, and 
$y=0$ denotes the absence of wildfire risk on the prediction day. This binary classification setting is commonly used in operational wildfire forecasting systems, where timely and accurate predictions can support early warning efforts and inform mitigation strategies. By tracking changes in temperature, humidity, wind, and precipitation over time, AI models identify patterns that may warn of a higher risk of wildfires.  Incorporating time series structure allows the model to consider the current environmental state and to account for trends, lags, and cumulative effects that develop over the preceding days or weeks.

\subsection{Model Explanations and SHAP}

The goal of model explanations is to improve transparency, interpretability, and trust in ML  models by  capturing how each feature influences the model's decisions and which class such decisions favor.  XAI methods address this by providing human-understandable reasons for model behavior. 
SHapley Additive exPlanations (SHAP) \cite{lundberg2017unified} is a widely used XAI method. Given a sample $x = \{x_i\}_{i=1}^N$ where $x_i$ represents the  $i^{th}$ feature and $N$ is the number of features, and  a prediction model $f$ that outputs the probability  $f(x)$ of $x$ belongs to a certain class $y$,  SHAP  can be represented as a linear model of the form $g(x) = \sum_{i=1}^N e_i x_i$, where $\{e_i\}_{i=1}^d$ are the resulting coefficients of the explanation model $g(x)$,   measuring the impact of   $x_i$ on the model's decision. 
In general, higher values of $e_{i}$ imply a higher impact of   $x_i$ in the model decision.

% \subsection{Data for Weather-Forecasting Tasks}
% The data types used for weather forecasting vary across different tasks and models. For instance, Convolutional Neural Networks and Vision Transformers can process reanalysis data, which are currently the most complete ... of past weather conditions. Reanalysis combines all distinct weather observations available on one given day in the past, together with physical laws introduced by a sophisticated weather computer model, to produce an image representing weather conditions on that specific day (\cite{copernicus2020}). On the other hand, LSTM models and other Transformer-based architectures are more suited to handle multivariate time series, with each variable clearly defined and having its own range of values. 

\section{Transparent Wildfire Modeling: Insights through XAI }

In this section, we shed light on how XAI can enhance our understanding of wildfire predictions and support preparedness activities. To achieve this, we  examine different AI models and use SHAP  with an adjusted visualization to interpret the outputs of these models. Our analysis reveals 1) how different AI models make decisions regarding wildfire risk, 2) which features drive predictions, and 3) how insights into these contributing factors can inform and improve preparedness strategies. 
By revealing the internal reasoning of AI models, XAI empowers decision-makers to move beyond black-box predictions toward transparent preparedness planning. The insights derived from this analysis can guide more effective strategies for wildfire mitigation and emergency management.

% This section explores how XAI enhances wildfire prediction and preparedness. We examine multiple AI models and use SHAP with adjusted visualization to interpret their outputs. Our analysis reveals (1) how models assess wildfire risk, (2) which features drive predictions, and (3) how these insights inform preparedness strategies. By opening the black box, XAI supports transparent, effective planning for wildfire mitigation and emergency management.

\subsection{AI Models for Wildfire Prediction}
To ensure generalizable analysis,  we employ various AI models widely used for wildfire prediction,  including \textit{deep learning models}, i.e.,  LSTM, Transformer, GTN, and \textit{tree-based models}, i.e., Random Forest and XGBoost. 
They capture complex temporal patterns  in wildfire data and allow comparison of  accuracy and interpretability across   architectures. 

% To ensure generalizable analysis, we use AI models common in wildfire prediction, including deep learning (LSTM, transformer, GTN) and tree-based methods (random forest, XGBoost). These capture complex temporal patterns and enable comparison of performance and interpretability across architectures.

LSTM is a recurrent neural network designed to capture long-term dependencies using gating mechanisms, making it effective for sequential data like wildfire events \cite{staudemeyer2019understanding}. Transformer models rely on self-attention rather than recurrence to model sequence dependencies; we adopt an encoder-only version for wildfire risk prediction \cite{vaswani2017attention}. GTN extends the Transformer by adding dual attention, making it well-suited for multivariate time-series data in environmental settings \cite{liu2021gated}.
We also include two ensemble tree-based models, including Random Forest and XGBoost, known for handling noisy and heterogeneous data. Random Forest uses multiple decision trees with majority voting to improve robustness \cite{breiman2001random}, while XGBoost builds trees sequentially to refine predictions, improving accuracy and efficiency.

We train deep learning models using the Adam optimizer with the following settings: LSTM with  a learning rate ($lr$) of $0.004$ and  a weight decay ($wc$) of $0.0063$, Transformer with $lr = 0.0001$ and $wc=0$, and GTN with $lr=0.0012$ and  $wc=0.0045$. All models run for $30$ epochs with a $0.1$ learning rate decay every $15$ epochs. 
% Each model was trained for $30$ epochs, with a learning rate decay factor of $0.1$ applied every $15$ epochs. 
We adopt model architectures and training pipelines from the open-source Mesogeos project \cite{kondylatos2023mesogeos}. 
For the tree-based models,  we use \verb|sklearn| implementations with $100$ trees each. Random Forest uses Gini impurity with a minimum split size of $2$ and no depth limit, while XGBoost uses a learning rate of 0.3 and a max depth of $6$. %These ensemble models required minimal tuning.

% the built-in implementations provided by the sklearn library, configuring each to use $100$ decision trees.  For Random Forest, we used Gini impurity as the criterion for splitting an internal node, with a minimum of $2$ samples required for a split and no maximum tree depth. In the case of XGBoost, we set the learning rate at $0.3$, with a maximum depth of $6$ for each tree. These ensemble models required minimal tuning and were used primarily to complement the deep learning approaches with interpretable baselines.

% \phung{add information about the learning rate, optimizer, weight decay information for decision tree and XGBoost.}  

\subsection{Dataset}
We conducted our analysis of wildfire prediction and XAI using two datasets from distinct geographical regions. The diverse approach allows us to examine how wildfire prediction may vary across different regions, particularly in terms of feature importance, and to draw broader, more generalizable insights for preparedness and mitigation.

\subsubsection{General Information}

The Mesogeos dataset is a large-scale dataset for wildfire modeling in the Mediterranean \cite{kondylatos2023mesogeos}. We use its wildfire danger forecasting subset, containing $25,722$ samples: $19,353$ for training, $2,262$ for validation, and $4,107$ for testing. Each sample includes $24$ features over $30$ days prior to a wildfire event, split into static (i.e., values remain the same for all days within a sample) and dynamic (i.e., values might change daily) types. Table~\ref{tab:mesogeos_vars} provides the full name and explanations for each feature in this dataset. 

% The first dataset, Mesogeos, is  a large-scale multi-purpose dataset for wildfire modeling in the Mediterranean \cite{kondylatos2023mesogeos}. For this study, we used the subset designated for wildfire danger forecasting, which includes $25,722$ samples. Among them, $19,353$ were used for training, $2,262$ for validation, and $4,107$ for testing. The training and validation samples were used to train the AI models, while the testing samples were used to produce the visualizations in this paper. Each sample comprises $24$ features collected over $30$ consecutive days preceding a wildfire event.  The features are divided into two classes, static and dynamic. The values of static features remain the same for all days within a sample, while the values of dynamic features might change daily.  
% Table \ref{tab:mesogeos_vars} provides the full name for each feature in this dataset. For the purposes of interpretability and consistency in XAI analysis, we produced our visualizations using only samples that were correctly classified by the models as wildfire occurrences.

The California Wildfires dataset covers weather conditions and wildfire occurrences in California from $1984$ to $2025$ \cite{yavas2025calfire}.  The dataset includes nine feature groups, with seasons one-hot encoded. Table~\ref{tab:calfire_vars} provides the full name for each feature in this dataset.   A total of $14,976$ observations are grouped into $1,248$ samples, each representing an $11$-day window preceding a wildfire event. Among them, $998$ samples are used for training and $250$ samples for testing.

When performing XAI analysis with each dataset and each AI model, we only used samples that were correctly predicted as wildfire occurrences by the model in question. This sample selection is used to ensure that the results presented in our final visualizations are both consistent and reliable.

% The dataset has 14976 observations in total, which are grouped into 1248 samples, with each sample including $11$ days of observations leading up to a wildfire event. The training set consists of $998$ samples, while the test set contains $250$ samples.

% The second dataset, referred to as California Wildfires, contains data on weather conditions and wildfire occurrences in the state of California, USA, from 1984 to 2025 \cite{yavas2025calfire}. For model training and evaluation, we excluded time indicators such as date, month, and year. The final dataset is composed of nine feature groups, with the season group one-hot encoded into binary categorical features. Table \ref{tab:calfire_vars} provides a complete description of the features used. \kv{The dataset has 14976 observations in total, which are grouped into 1248 samples,} with each sample including $11$ days of observations leading up to a wildfire event. The training set consists of $998$ samples, while the test set contains $250$ samples.

\begin{table*}[t] 
\centering 
\caption{Features in the Mesogeos dataset.} 
\begin{tabular}{|c|c|c|}
 \hline
\textbf{Features} & \textbf{Abbreviation} & \textbf{Full name} \\ 
  \hline
\multirow{5}{4em}{Static} & population, slope, dem, roads\_distance & Population, Slope, Elevation, Distance from roads \\
                    \cline{2-3}
& lc\_wetland, lc\_shrubland, lc\_grassland  & Area of wetland, shrubland, grassland \\
                    \cline{2-3}
& lc\_water\_bodies, lc\_forest & Area of water bodies, forest \\  
\cline{2-3}
& lc\_sparse\_vegetation & Area of sparse vegetation \\
\cline{2-3}
& lc\_settlement, lc\_agriculture  & Area of settlement land, agricultural land \\
\hline
\multirow{4}{4em}{Dynamic} & wind\_speed, ssrd & Wind speed, Surface solar radiation downwards\\
\cline{2-3}
  &   tp, sp, rh & Total precipitation, Surface pressure, Relative humidity \\
\cline{2-3}
   &  t2m, d2m, lst\_night, lst\_day  & 2-meter temperature, Dewpoint temp., Night's and  day's land surface temp. \\
\cline{2-3}
  &   smi, lai, ndvi & Soil moisture index, Leaf area index, Normalized difference vegetation index \\
\hline                 
\end{tabular}
\label{tab:mesogeos_vars}  
\vspace{-2mm}
\end{table*}

\begin{table*}[t] 
\centering 
\caption{Features in the California Wildfires dataset. } 
\begin{tabular}{|c|c|}
 \hline
\textbf{Name} & \textbf{Explanation} \\ 
  \hline
precipitation, lagged\_precipitation  & Daily precipitation, Cumulative precipitation over the preceding 7 days\\
                    \hline
max\_temp, min\_temp, temp\_range & Maximum daily temp., Minimum daily temp., Daily temperature range\\
\hline
avg\_wind\_speed, wind\_temp\_ratio, lagged\_avg\_wind\_speed  & Average daily wind speed, Ratio of average wind speed to max temp.\\
\hline
lagged\_avg\_wind\_speed & Average wind speed over preceding 7 days \\
\hline
season & The season of the observation (Winter, Spring, Summer, Fall) \\
\hline                 
\end{tabular}
\label{tab:calfire_vars}  
\vspace{-2mm}
\end{table*}

\subsubsection{Exploratory Data Analysis}

\begin{figure*}[t]
 \centering
\subfigure[lst\_day]{\includegraphics[scale=0.29]{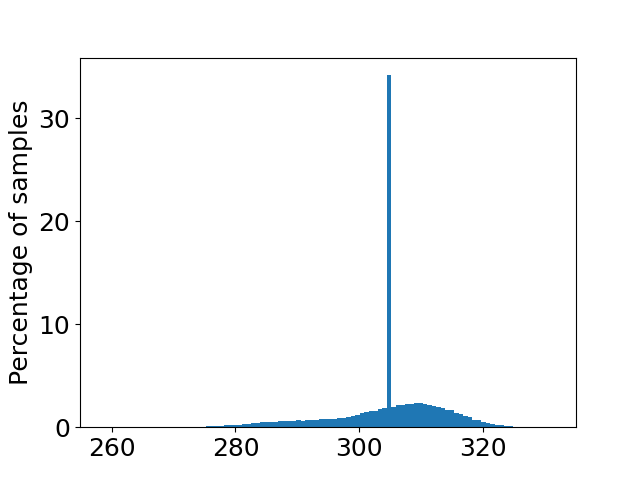}} \hspace{0.35cm}
\subfigure[ndvi]{\includegraphics[scale=0.29]{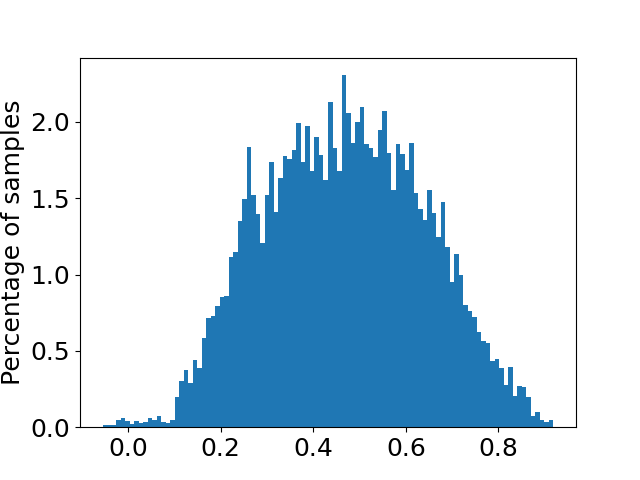}} \hspace{0.35cm}
\subfigure[population]{\includegraphics[scale=0.29]{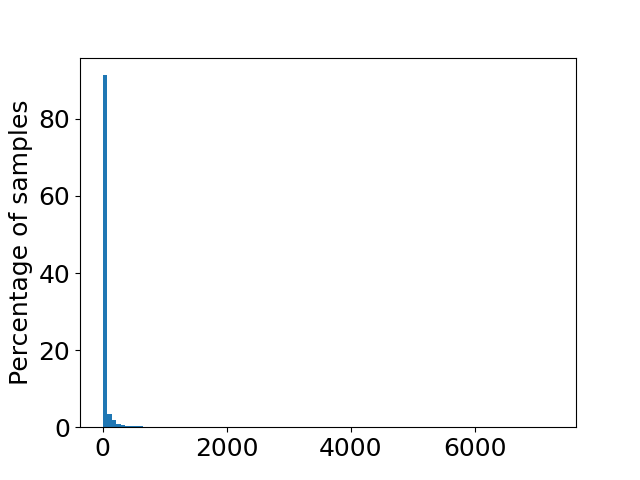}}  
\caption{Representative feature distributions from Mesogeos dataset.} 
\label{fig:mesogeos_dist}
\end{figure*}

% \phung{add Mesogeos data information here} 
\textit{In the Mesogeos dataset}, six features have missing values across multiple samples. Table \ref{tab:mesogeos_missing} presents their average percentage of missing values, with \verb|lst_day| and \verb|lst_night| missing in about one-third of the total entries. The original   preprocessing pipeline filled missing values with zeros \cite{kondylatos2023mesogeos}, but for temperature features in Kelvin, this implies absolute zero, which is a physically unrealistic   condition.  %temperature features are measured in Kelvin, filling missing temperatures with zeros would imply absolute zero, a physically unrealistic and uninhabitable condition.
 To address this, we impute missing values using the mean of each feature  calculated from the training set.  While this method introduces noticeable skewness, particularly in \verb|lst_day| and \verb|lst_night|, mean imputation is more physically plausible than zero-filling and does not significantly degrade data utility. In fact, random checks with the LSTM, Transformer, and GTN models show higher accuracy using mean imputation compared to zero-filling. 
 Another feature with similar skewness is \verb|smi|, with $6.83\%$ of its values imputed. 
 % Other dynamic features like \verb|ndvi| and \verb|rh| show roughly Gaussian distributions with milder skewness. Static features, such as land cover and population, mostly have values near zero, reflecting the Mediterranean region's large water bodies and limited landmass. Examples of these distributions are shown in Figure \ref{fig:mesogeos_dist}.
 Other dynamic features (e.g., \verb|ndvi|, \verb|rh|) follow near-Gaussian distributions, while static features (e.g., land cover, population) cluster near zero, reflecting the Mediterranean landscape (Figure \ref{fig:mesogeos_dist}).

\begin{table*}[t] 
\centering 
\caption{Percentage of missing values on several features in the Mesogeos dataset.} 
\begin{tabular}{|c|c|c|c|c|c|c|c|c|c|c|}
 \hline
% \textbf{Variable} & \textbf{Average percentage of missing values (\%)} \\
%   \hline
\textbf{Feature} & lst\_night & lst\_day &smi &lai &ndvi &population \\
\hline 
\textbf{Percentage (\%)} & $36.43$  & $31.49$  & $6.83$ &$1.69$ & $0.25$ & $0.01$ \\
\hline             
\end{tabular}
\label{tab:mesogeos_missing}  
\vspace{-2mm}
\end{table*}

\begin{figure*}[t]
 \centering
\subfigure[max\_temp]{\includegraphics[scale=0.28]{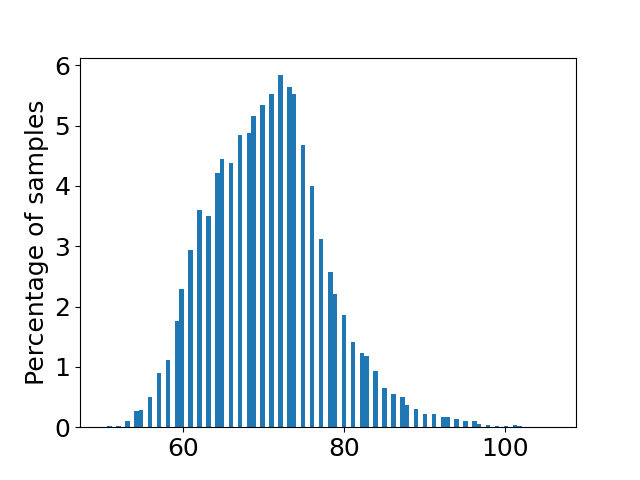}} \hspace{0.35cm}
\subfigure[min\_temp]{\includegraphics[scale=0.28]{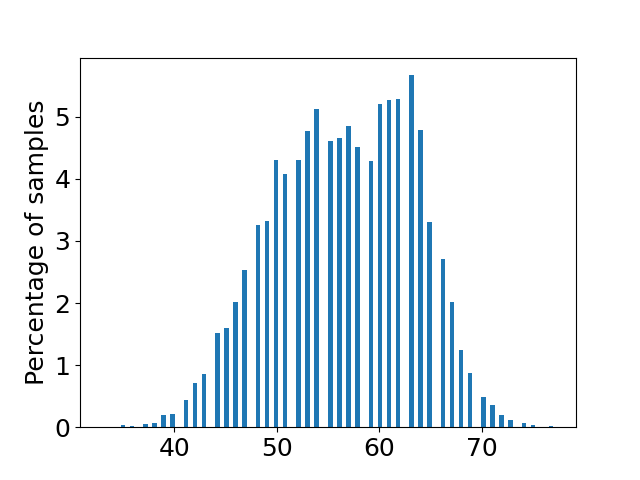}} \hspace{0.35cm}
\subfigure[precipitation]{\includegraphics[scale=0.28]{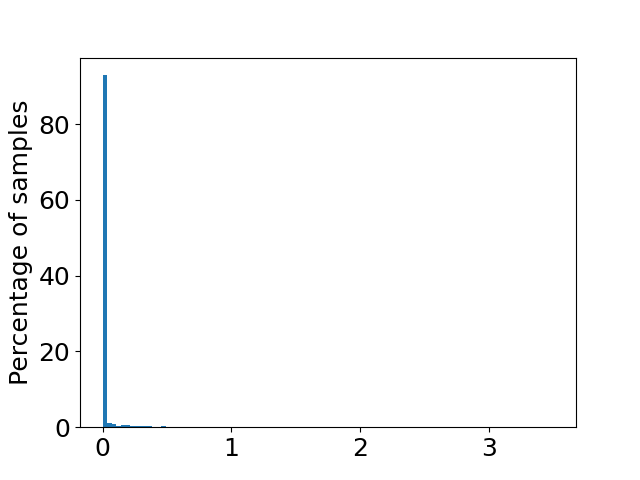}}   
\caption{Representative feature distributions from California Wildfires dataset.} 
\label{fig:calfires_dist}
\end{figure*}

\textit{In the California Wildfires dataset}, there are a total of $11$ features, comprising $8$ original features and a one-hot encoded \verb|season| feature represented by three mutually exclusive binary variables: \verb|season_winter|, \verb|season_spring|, and \verb|season_summer|. 
% nine original features. Among them, the season feature is represented by three mutually exclusive binary categorical features, including season\_winter, season\_spring, season\_summer, which are one-hot encoded. Therefore, the total number of features is $11$. 
Features such as precipitation, maximum temperature (\verb|max_temp|), minimum temperature (\verb|min_temp|), and average wind speed (\verb|avg_wind_speed|) are treated as independent. Other features are derived or correlated, such as \verb|temp_range| (from \verb|max_temp| and \verb|min_temp|) and \verb|wind_temp_ratio| (from \verb|avg_wind_speed| and \verb|max_temp|). Lagged features are computed over a 7-day window. The seasonality variables are mutually exclusive: if \verb|season_summer| is $1$, then \verb|season_winter| and \verb|season_spring| are $0$, and vice versa. This dataset has one day with missing values in \verb|precipitation|, \verb|min_temp|, \verb|max_temp|, and \verb|temp_range|. Additionally, there are 12 missing entries in \verb|avg_wind_speed|, which also affect \verb|wind_temp_ratio|. We impute missing values in the independent features using their global means. Missing \verb|temp_range| and \verb|wind_temp_ratio| values are calculated using data from their defining features. Looking deeper into the dataset, the independent features \verb|min_temp|, \verb|max_temp|, and \verb|avg_wind_speed| exhibit roughly Gaussian distributions. Both \verb|max_temp| and \verb|avg_wind_speed| are right-skewed, while \verb|min_temp| is slightly left-skewed and bimodal, reflecting seasonal temperature shifts. The remaining indepedent feature, \verb|precipitation|, is highly skewed, with $90.78\%$ of values equal to zero, producing a near-uniform distribution mirrored in its lagged version. Derived features such as \verb|temp_range|, \verb|wind_temp_ratio|, and \verb|lagged_avg_wind_speed| are also right-skewed. Seasonality features are uniformly distributed, with each binary variable set to $1$ in roughly $25\%$ of samples. Figure~\ref{fig:calfires_dist} shows representative distributions in this dataset.

\vspace{-2mm}
\subsection{Visualizing the Temporal Evolution of SHAP Values}
% In this section, we introduce \textsc{XClimate}, an XAI framework that utilizes model explanations to visualize the impacts of variables on the model's decisions. 
% This framework operates entirely in a blackbox setting and extends existing XAI framework, such as SHAP, making it easy to use and cost-efficient. 

In this study, we leverage the SHAP framework \cite{lundberg2017unified} to explain wildfire predictions. While default SHAP visualizations, such as bar or summary plots, are effective for static data, they are inadequate to  capture the temporal dynamics of time series. Specifically, they do not show how the influence of individual features evolves over time, which is crucial for understanding sequential patterns leading to wildfire events and identifying potential forecast opportunities. %, and potentially indicates forecast opportunities for these events. 

To address this limitation, we extract SHAP values for each feature across multiple time steps and create a custom scatterplot visualization. For each feature, we plot its SHAP values over the days leading up to a wildfire event, such as $30$ days for the Mesogeos dataset and $11$ days for the California Wildfires dataset. In this plot, dot color and size represent the direction and magnitude of the feature's impact, respectively. Using a blue-white-red colormap, blue dots indicate negative SHAP values, white represents near zero, and red denotes positive values. Dot size is proportional to the absolute SHAP value, emphasizing more influential contributions. This visualization provides a clearer and more intuitive view of how key features affect model decisions over time, supporting more informed and actionable wildfire preparedness.

\begin{table*}[t] 
\centering 
\caption{Accuracy (\%) of models trained on Mesogeos and California Wildfires datasets.} 
\begin{tabular}{|c|c|c|c|c|c|c|}
 \hline
\multicolumn{2}{|c|}{Model} & LSTM & Transformer & GTN & Random Forest & XGBoost \\
\hline
\multirow{2}{3.5em}{Accuracy} &  Mesogeos & 87.00 & 87.53 & 86.34 & 77.23 & 75.00\\ 
\cline{2-7}
&  California Wildfires & 76.80 & 78.71 & 77.60 & 76.31  & 71.89 \\
\hline    
\end{tabular}
\label{tab:models}  
\end{table*}

% \begin{table}[t] 
% \centering 
% \caption{Accuracy of models trained on Mesogeos and California Wildfires datasets.} 
% \begin{tabular}{|c|c|c|}
%  \hline
% \multirow{2}{6em}{Model's name} & \multicolumn{2}{c|}{Accuracy (\%)} \\ 
% \cline{2-3}
% & \textbf{Mesogeos} & \textbf{California Wildfires} \\
%   \hline
% LSTM & 87.00 & 76.80 \\
% \hline
% Transformer & 87.53 & 78.71 \\
% \hline
% GTN & 86.34 & 77.60 \\
% \hline
% Random Forest & 77.23 & 76.31 \\
% \hline
% XGBoost & 75.00 & 71.89 \\
% \hline    
% \end{tabular}
% \label{tab:models}  
% \end{table}

% \begin{table}[t] 
% \centering 
% \caption{Accuracy of models trained on Mesogeos and California Wildfires datasets.} 
% \begin{tabular}{|c|c|c|c|c|}
%  \hline
% \multicolumn{5}{|c|}{\textbf{Mesogeos}} \\ 
%   \hline
% Name & Accuracy (\%) & Precision & Recall & F1 Score \\
% \hline
% LSTM & 87.00 & 0.79 & 0.84 & 0.81 \\
% \hline
% Transformer & 87.53 & 0.78 & 0.87 & 0.82 \\
% \hline
% GTN & 86.34 & 0.77 & 0.85 & 0.81 \\
% \hline
% Random Forest & 77.23 & 0.68 & 0.59 & 0.64 \\
% \hline
% XGBoost & 75.00 & 0.62 & 0.66 & 0.64\\
% \hline   
% \multicolumn{5}{|c|}{\textbf{California Wildfires}} \\ 
%   \hline
% Name & Accuracy (\%) & Precision & Recall & F1 Score\\
% \hline
% LSTM & 76.80 & 0.64 & 0.62 & 0.63 \\
% \hline 
% Transformer & 78.71 & 0.63 & 0.66 & 0.65 \\
% \hline
% GTN & 77.60 & 0.70 & 0.57 & 0.63 \\
% \hline
% Random Forest & 76.31 & 0.63 & 0.47 & 0.53 \\
% \hline
% XGBoost & 71.89 & 0.57 & 0.37 & 0.45 \\
% \hline    
% \end{tabular}
% \label{tab:models}  
% \end{table}

\vspace{-2mm}
\subsection{Experimental Results}

\subsubsection{Prediction Performance }
% Table \ref{tab:models} shows the prediction accuracy  across different models and datasets. Overall, Random Forest and XGBoost maintain relatively similar performances for both datasets, while LSTM, Transformer, and GTN perform much better on Mesogeos. The biggest difference is in LSTM's performance across the two datasets, with a gap of 13.51 percentage points. This discrepancy could be due to Mesogeos having significantly more training samples, which allows the models to learn more from hidden patterns in the data.

Table \ref{tab:models} presents the prediction accuracy of five AI models evaluated on the Mesogeos and California Wildfires datasets. Deep learning models consistently outperform tree-based ensemble models across both datasets. On the Mesogeos dataset, the Transformer achieves the highest accuracy at 87.53\%, followed closely by LSTM (87.00\%) and GTN (86.34\%). In contrast, Random Forest and XGBoost perform worse, with accuracies of 77.23\% and 75.00\%, respectively.   We observe a similar trend on the California Wildfires dataset, although the overall accuracies are lower due to its smaller size and higher variability. The Transformer again leads with 78.71\% accuracy, followed by GTN (77.60\%) and LSTM (73.49\%), while Random Forest and XGBoost achieve 76.31\% and 71.89\%, respectively. These results highlight the strength of deep learning models, particularly those based on attention mechanisms, in capturing the complex temporal patterns associated with wildfire prediction.

\subsubsection{SHAP Explanations and Visualization} \label{exp_viz}

In this section, we enhance interpretability by visualizing SHAP values over time, showing how features influence wildfire predictions across sequential steps. We apply this to multiple AI models and both datasets, offering general insights here and detailed analysis in Section \ref{sec:Discussion}. 
% Key visualizations appear in the main paper, with extra figures in the supplement.

% In this section, we enhance the interpretability of model predictions by visualizing SHAP values across time. This approach enables a more intuitive understanding of how different features contribute to wildfire predictions over sequential time steps. We apply this technique to multiple AI models and analyze their behavior across both datasets. While we offer general observations here, we present a more in-depth analysis in Section \ref{sec:Discussion}. For clarity and focus, we include only the most insightful visualizations in the main paper and include additional figures in the supplementary materials.

% In this section, we present visualizations of SHAP values for the models across both datasets, as well as some general commentary on the information within those visualizations; we provide more in-depth analysis in the Discussions section. Overall, the patterns in SHAP values for different models learning from the Mesogeos dataset are largely similar, while the patterns observed across models trained on the California Wildfires dataset exhibit some differences. To highlight noteworthy observations and to better convey our points, we opted to show only certain interesting visualizations in the paper; the remaining figures have been included in the supplementary materials. 

\begin{figure*}[t]
      \centering %\{-10pt}
       \includegraphics[scale=0.15]{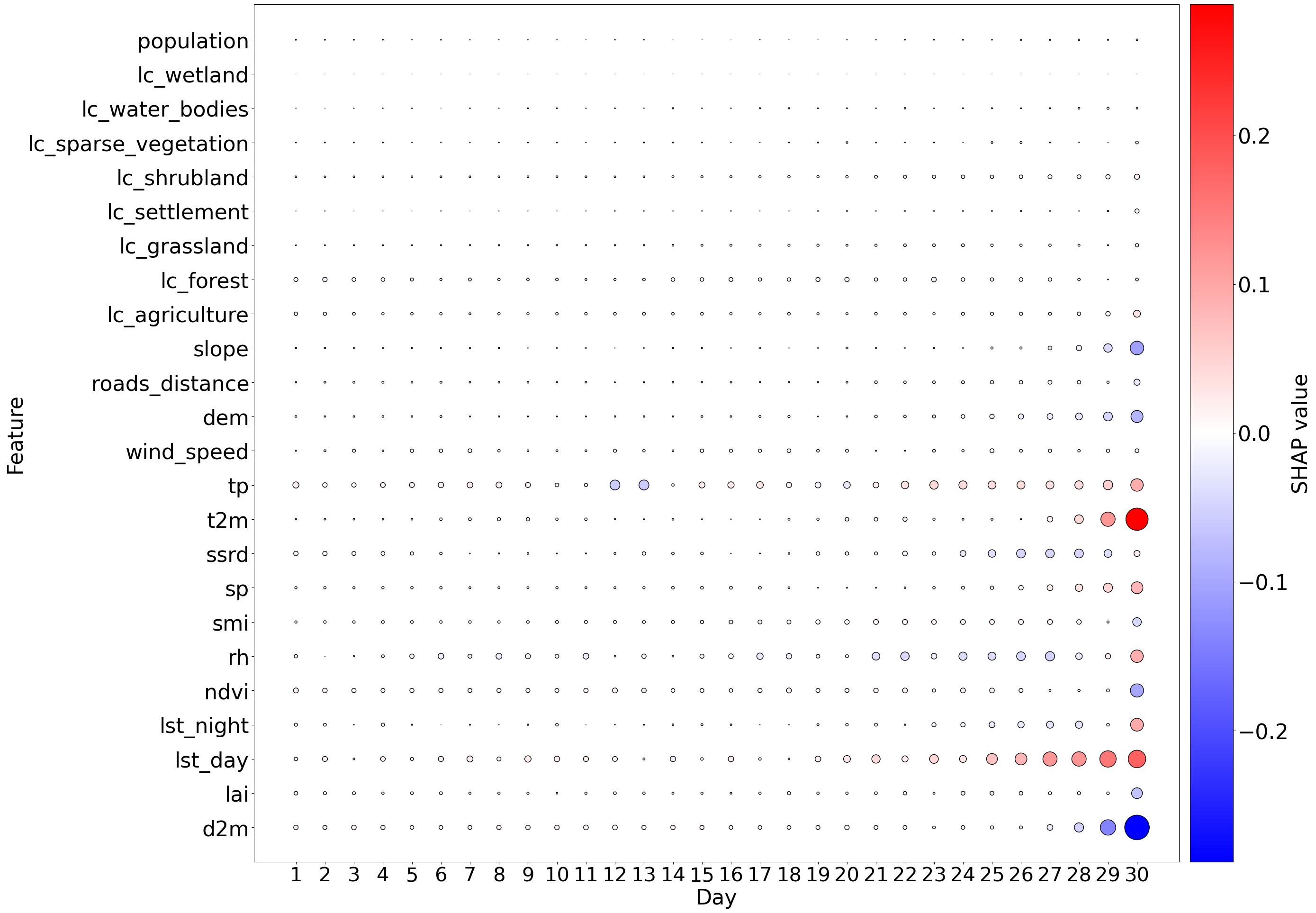} 
      \caption{Visualization for \textit{average} SHAP values for the Mesogeos dataset using the LSTM model.}
      \label{fig:mesogeos_lstm_avg}
\vspace{-2mm}
\end{figure*}

\begin{figure*}[t]
      \centering %\{-10pt}
       \includegraphics[scale=0.24]{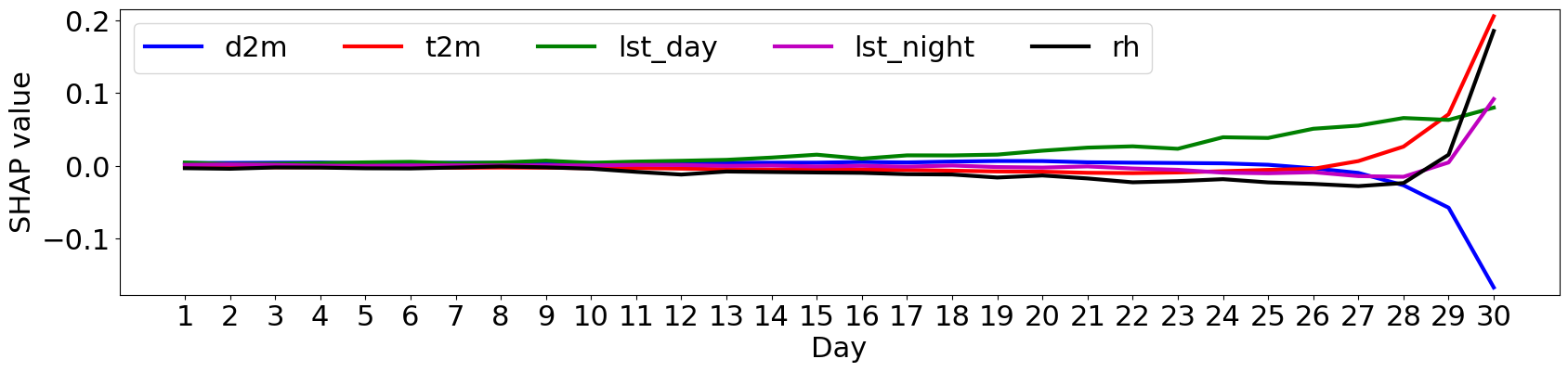} 
      \caption{SHAP value evolution in the top five important features    (Mesogeos and LSTM model).}
      \label{fig:curve}
\vspace{-2mm}
\end{figure*}

\begin{figure*}[t]
    \centering
    \subfigure[Mesogeos]{\includegraphics[scale=0.15]{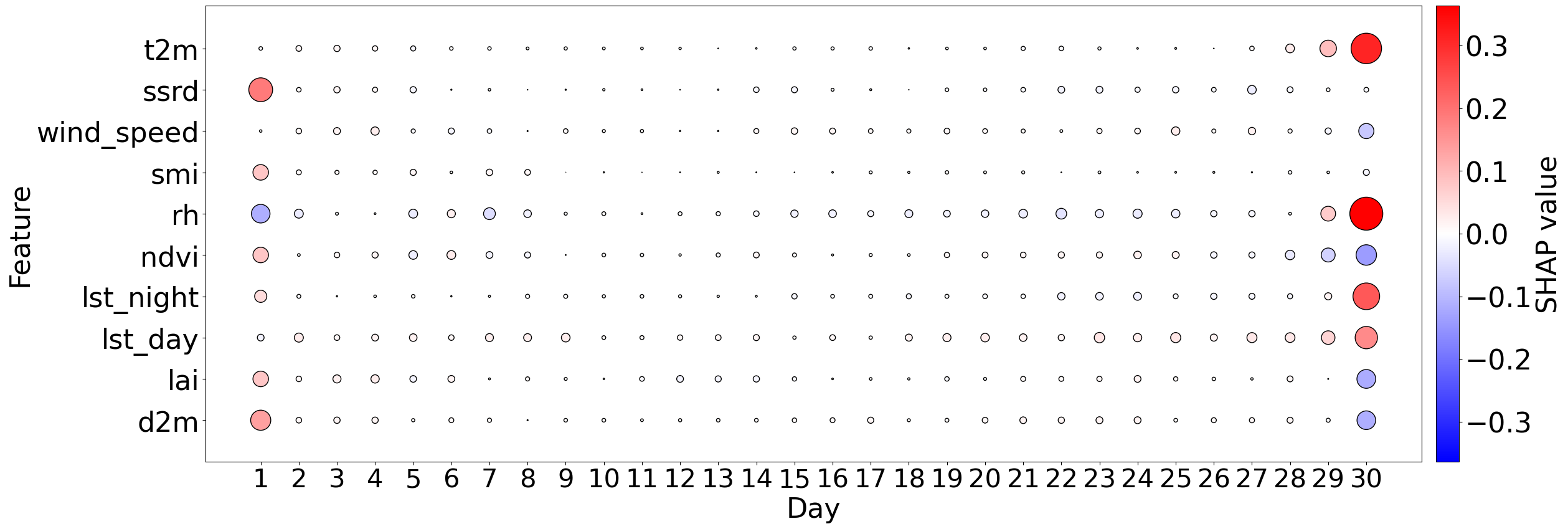}} \hfill
    \subfigure[California Wildfires]{\includegraphics[scale=0.17]{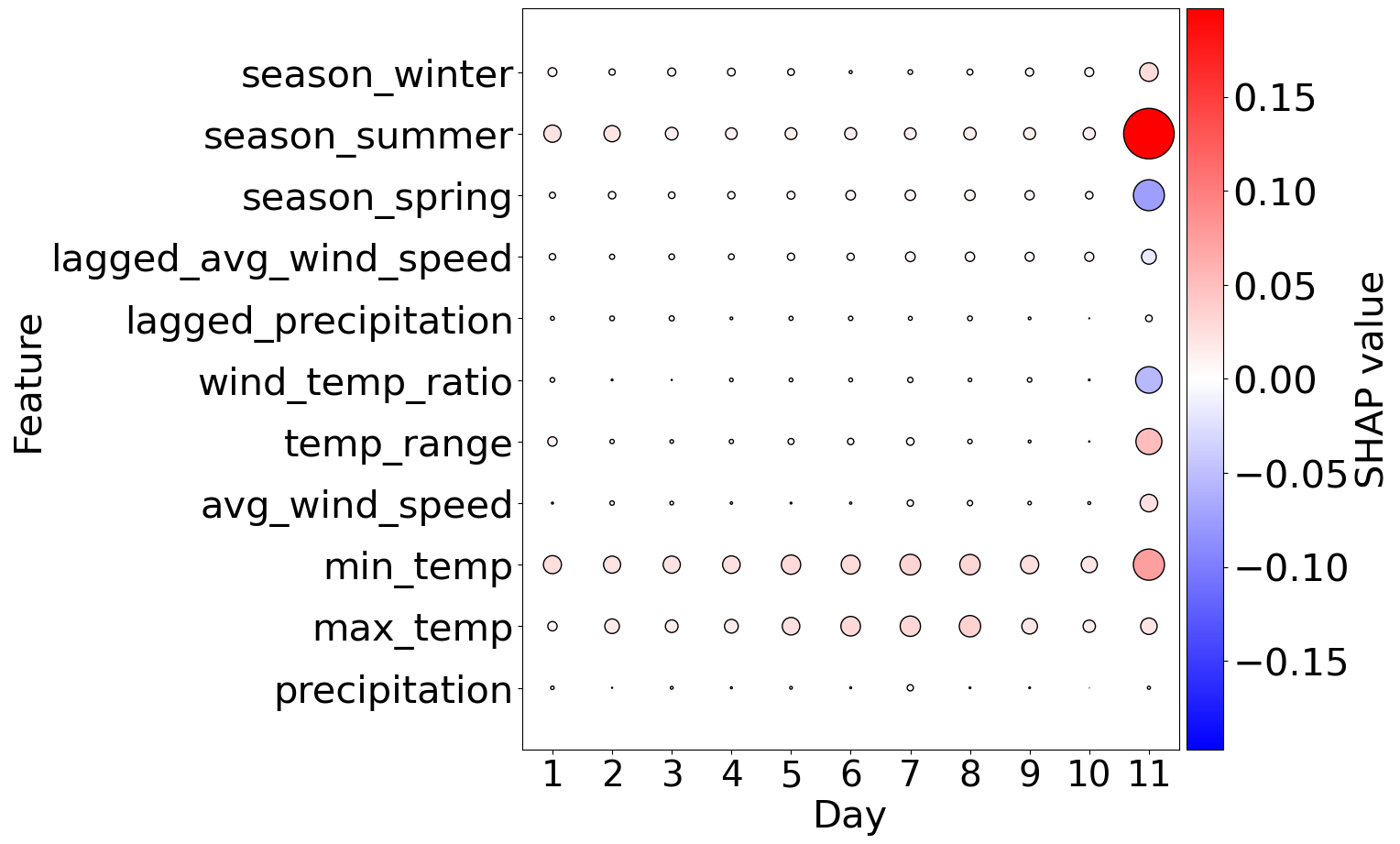}} \hfill
    \caption{Average SHAP values in the Transformer model across datasets.} 
    \label{fig:transformer_diff}
\vspace{-2mm}
\end{figure*}

\textbf{Mesogeos Dataset.} Figure \ref{fig:mesogeos_lstm_avg} presents the average SHAP values across all   samples in the Mesogeos dataset using an LSTM prediction model. Different from other visualizations \cite{fan2024explainable, cilli2022explainable}, we display all $24$ features over a $30$-day window. Overall, key features driving the model’s wildfire predictions include 2-meter air temperature (\verb|t2m|), relative humidity (\verb|rh|), daytime and nighttime land surface temperature (\verb|lst_day|, \verb|lst_night|), and total precipitation (\verb|tp|). Conversely, dewpoint temperature (\verb|d2m|), soil moisture index (\verb|smi|), and elevation (\verb|dem|) negatively impact predictions, indicating lower wildfire likelihood. Other features, such as land cover classes, had minimal influence and are excluded from later figures to focus on more significant contributors. 
As expected, data closer to the prediction date have greater impact; for instance, when predicting day 31, weather data from day 25 onward contribute most. Figure \ref{fig:curve} shows how the importance of the top five features evolves over time, with notable changes beginning around day 28, matching the increased SHAP magnitudes seen in Figure \ref{fig:mesogeos_lstm_avg}. 
Feature importance also varies across individual cases, suggesting the model's reliance on certain features depends on temporal and environmental context. While some features show strong average effects, their impact can shifts based on local conditions, highlighting the model's dynamic behavior and the value of SHAP for uncovering such nuances.  Furthermore, the explanations generated by the Transformer-based models are consistent with the aforementioned observations with LSTM (Figure \ref{fig:transformer_diff}).

\begin{figure*}[t]
 \centering
\subfigure[Mesogeos]{\includegraphics[scale=0.15]{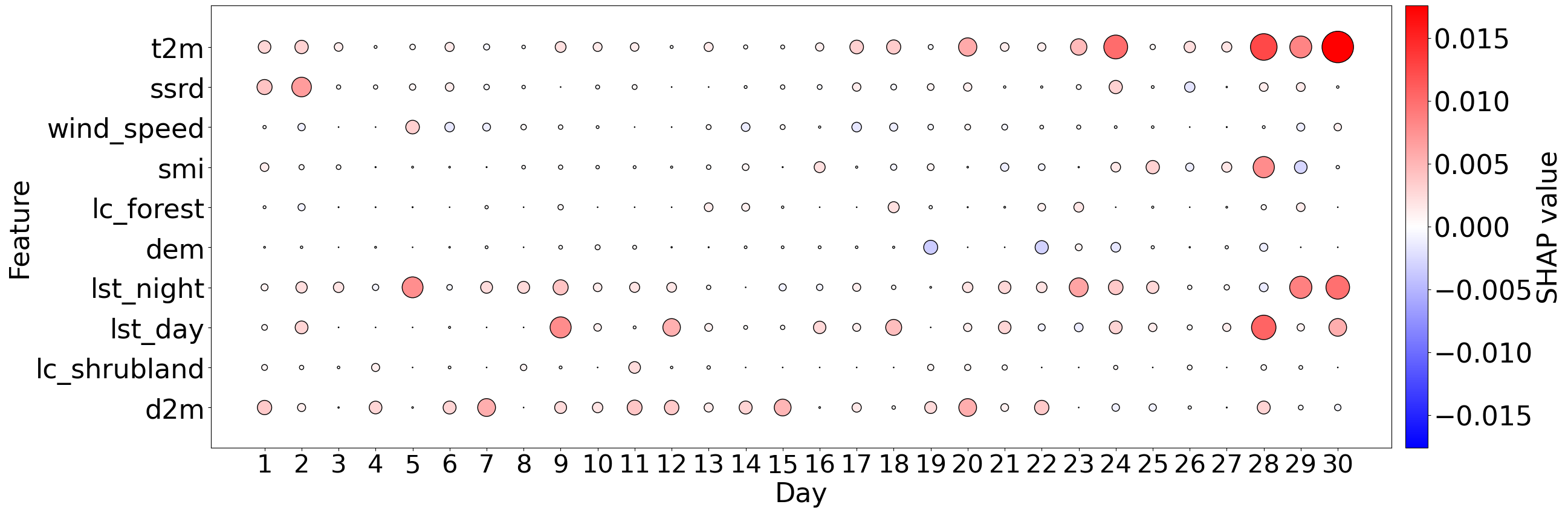}} \hfill
\subfigure[California Wildfires]{\includegraphics[scale=0.17]{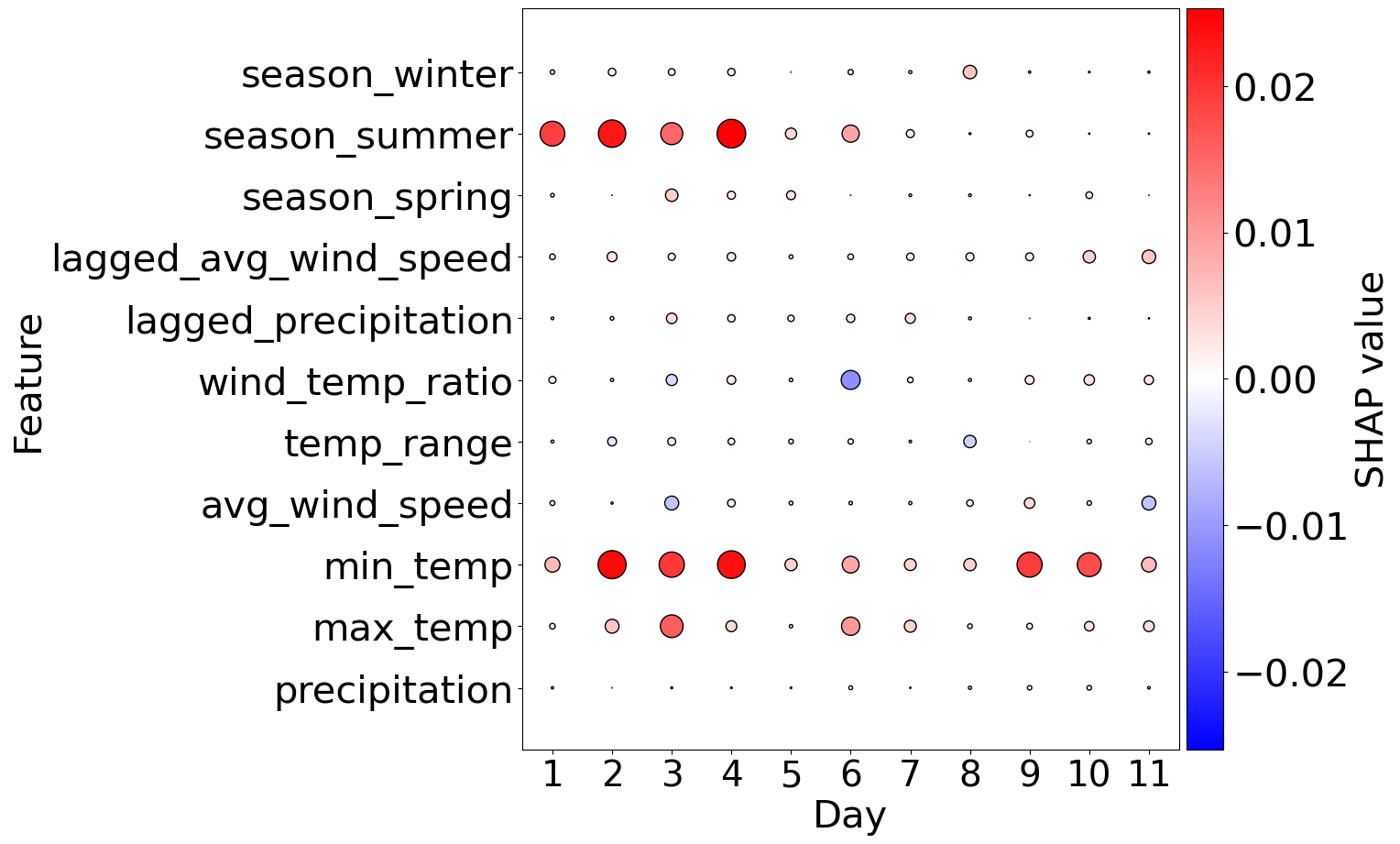}} \hfill
\caption{Average SHAP values in the Random Forest model across datasets.} 
\label{fig:rf_diff}
\vspace{-2mm}
\end{figure*}

In tree-based methods such as Random Forest and XGBoost, the evolution of feature contributions over the 30-day window differs from that of the deep learning models (i.e., LSTM, Transformer, GTN) when predicting wildfires on day $31$.  In Figure \ref{fig:rf_diff}, contributions come not only from the later days (e.g., day 25 onward) but also from earlier days, such as day 2 in \verb|ssrd| or day 5 in \verb|lst_night|. The root cause is that the Random Forest and XGBoost models do not use every feature at every time step for   node splits. As a result, some features may receive SHAP values of zero on certain days. Despite these   model architecture and temporal differences, the key features with the highest SHAP values remain largely consistent. In particular, temperature-related features continue to show strong correlations with wildfire risk across all models.

\textbf{California Wildfires Dataset.} 
We observe similar patterns in the California Wildfires dataset, where later days in the time window have a greater influence on wildfire predictions compared to earlier days. In addition, explanations from deep learning models (Figure \ref{fig:transformer_diff}b) slightly differ  from those of tree-based models (Figure \ref{fig:rf_diff}b).   
Similar to the Mesogeos dataset, temperature-related features also have the strongest impact on the model's predictions. SHAP values from individual samples show evolution consistent with the average SHAP values across the test set. Furthermore, explanations indicate that spring generally contributes the least to wildfire predictions, meaning a lower wildfire risk during this season.

\vspace{-2mm}
\subsection{Discussions}\label{sec:Discussion}
\subsubsection{Wildfire Prediction across Datasets and AI Models}

% Across datasets and AI models, we observe several common observations, including temperature-related variables had the most significant contributions to the wildfire prediction. In the Mesogeos dataset, relative humidity is another variable with high impact on a model's decision. In the California Wildfires dataset, seasonality has high impact to the prediction. In addition, the temporal evolution towards the wildfire prediction slightly differ between deep learning  and tree-based models, in which deep learning models show the later days have high impacts, and tree-based models show not only the later days but also several earlier days have impacts as well. This observation suggests that tree-based models like Random Forest and tree-based XGBoost may not be sufficiently  capture the role that time plays in wildfire prediction, even if these models can identify important variables in a manner consistent with other deep learning models. 

Across both datasets and all AI models, we observe several consistent patterns. First, temperature-related features contribute most significantly to wildfire predictions across all cases. In the Mesogeos dataset, relative humidity also plays a major role in model decisions, while in the California Wildfires dataset, seasonality emerges as a key influencing factor. Second, the temporal evolution of feature importance differs slightly between deep learning and tree-based models, and among deep learning models themselves. Whereas deep learning models tend to assign greater importance to later days, tree-based models rely on both later and some earlier days. This suggests that tree-based models such as Random Forest and XGBoost may be less effective in capturing temporal dynamics, even if they identify important features consistent with those found by deep learning approaches. 

In addition, different from LSTM, the Transformer model can extract meaningful signals from both the start and end of the input sequence. In Figure \ref{fig:transformer_diff}a,  features  such as \verb|ndvi|, \verb|lst_night|, and \verb|d2m| in the Mesogeos dataset  exhibit relatively high absolute SHAP values on both the first day and last day of the time window. Other variables, such as \verb|ssrd| and \verb|smi|, have lower SHAP values by the last day, but retain their overall positive contributions to model's decisions. This pattern is less pronounced in models trained on the California Wildfires dataset, likely due to its shorter time window and fewer samples. Consequently, Transformers trained on California data may lack the temporal depth to learn patterns similar to those from Mesogeos, resulting in different  trends.

\subsubsection{Seasonality of wildfires}
\textit{In the California Wildfires dataset}, SHAP explanations indicate that seasonality plays a significant role in predicting wildfire occurrence. Across AI models,  at least one season-related feature has a high SHAP value, with \verb|season_summer| being the strongest in Transformer and Random Forest models.  For the LSTM and GTN models, \verb|season_summer| and \verb|season_winter| have comparable impacts on model decisions. These results suggest a general consensus among the models that summer is the most fire-prone season, while certain winter conditions may also contribute to wildfire risk.  However, California's diverse climate, with wide temperature and precipitation variation across regions and years \cite{null2010weather}.   Further investigation into this geographic and climatic variability can clarify  seasonal  wildfire patterns   and improve model interpretations. %refine model interpretations of seasonality-related risk.

% Visualizations for SHAP values from the California Wildfires dataset show that seasonality has a significant impact on prediction of wildfire occurrence. With the exception of the XGBoost model, for all remaining models, at least one seasonality variable has a relatively large absolute SHAP value. In Transformer and Random Forest, season\_summer has the largest absolute SHAP value across all seasonality variables; for LSTM and GTN, season\_summer are more similar to season\_winter in its influence on the models' decisions. These observations suggest that the AI models we trained on the California Wildfires generally agree that summer days are most likely to witness wildfires among all seasons, while certain weather patterns in winter may also be conducive to fires. However, California's climate is very diverse, with mean temperature and precipitation often varying significantly across different areas of the state on a yearly basis \cite{enwiki:1292045240, null2010weather}. With further studies on this diversity, we may understand better certain trends in prediction of wildfires in California, such as the impact of each season.

\begin{figure*}[t]
 \centering
\subfigure[February]{\includegraphics[scale=0.18]{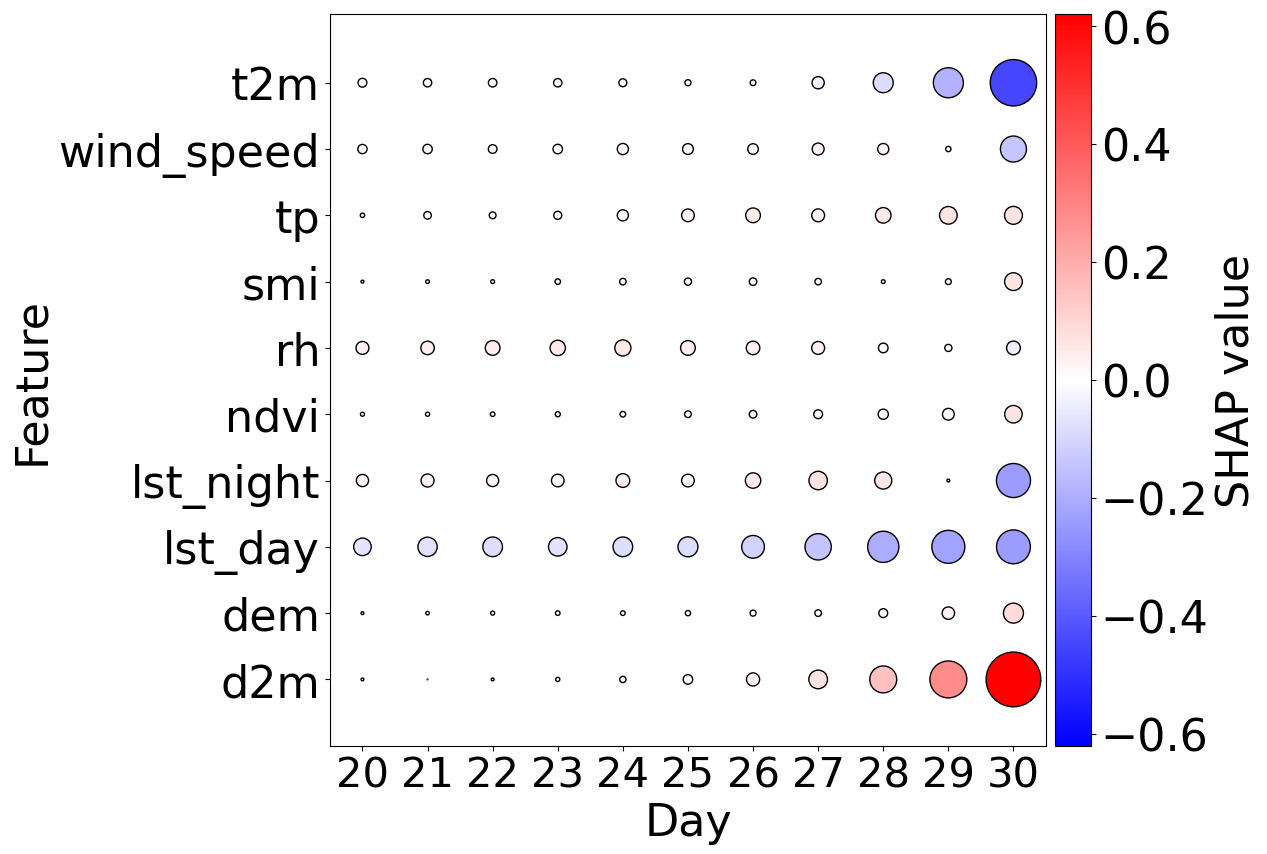}}\hspace{1cm}
\subfigure[July]{\includegraphics[scale=0.18]{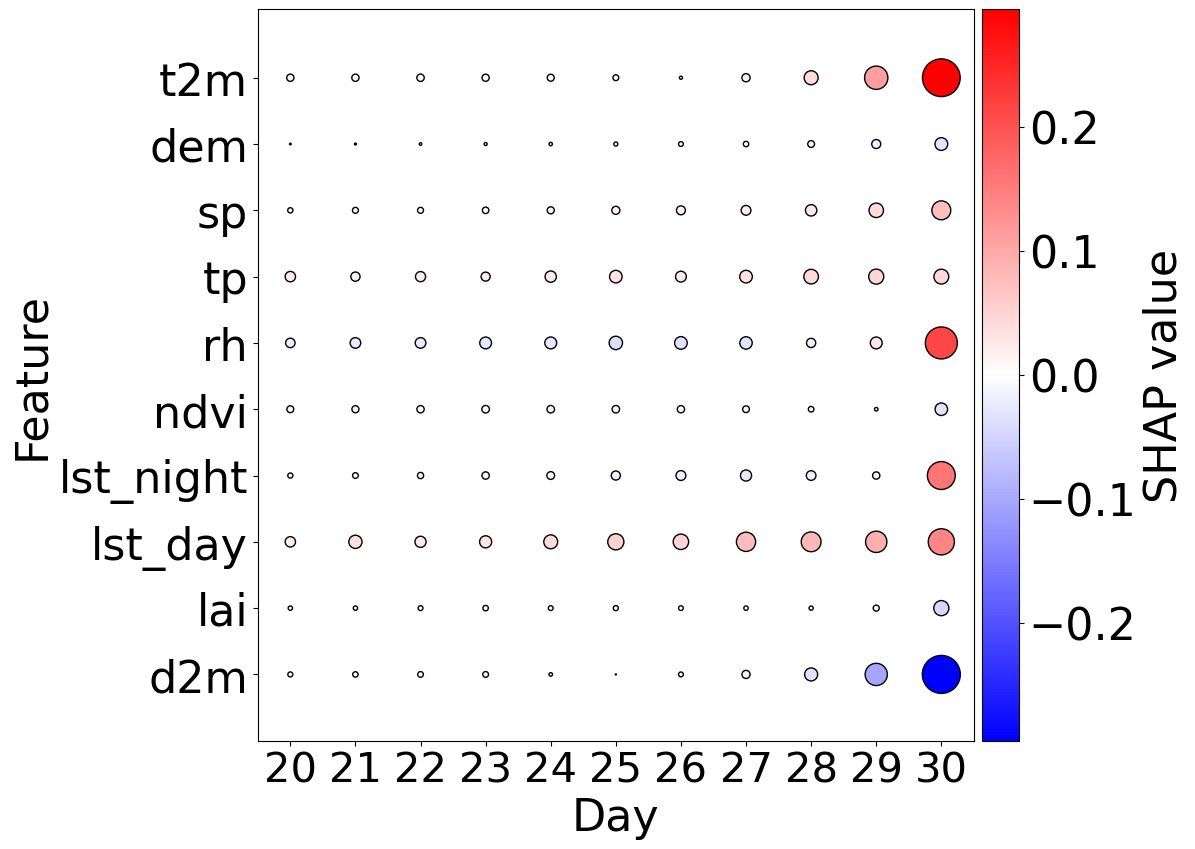}}\hfill
\caption{ Average SHAP values for wildfires in February and July (Mesogeos, LSTM model).} 
\label{fig:feb_july}
\end{figure*}

\textit{In the Mesogeos dataset}, no features explicitly encode seasonality. To examine seasonal effects, we visualize average SHAP values by month (Figure \ref{fig:feb_july}). From April to August, which are typically considered the summer period, temperature-related features strongly contribute positively, while dewpoint temperature contributes negatively, , which is consistent with the known physical variables that contribute to wildfires.  SHAP patterns vary by month, suggesting that wildfire risk factors change seasonally. Intuitively, higher temperatures and stronger winds increase wildfire risk, especially in summer, while more precipitation and higher dewpoint reduce it.
% higher temperatures and stronger wind speeds increase wildfire risk, especially during summer, whereas higher precipitation and dewpoint temperature tend to reduce it.  
During colder months, conditions affect fires differently.
Our results show that precursors and forecasts vary by region and season, highlighting the need to incorporate temporal context in wildfire prediction and interpretation.

\subsubsection{Forecast opportunities beyond weather time scale}

\begin{table*}[t] 
\centering 
\caption{Accuracy  and average training time  of Transformer model trained on different combinations of Mesogeos features.} 
\begin{tabular}{|c|c|c|c|c|}
 \hline
Feature combination & Top-5   & Top-10   & Top-20   & Original (24 features) \\
\hline
Accuracy (\%) (Most important) &  80.99 & 83.86 & 86.83 & 87.16\\ 
\hline   
% Accuracy (Absolute value) &  80.33 & 84.75 & 86.34 & 87.16\\ 
% \hline
Accuracy (\%) (Least important) &  73.90 & 80.11 & 85.63 & 87.16\\ 
% \hline
% Accuracy (\%) (Least important) &  82.18 & 82.85 & 84.84 & 87.16\\ 
% Accuracy (\%) (Least important; class 1) &  52.75 & - & - & -\\ 
% \hline
\hline
Training time per epoch (seconds) &  34.22 & 37.72 & 40.14 & 41.58 \\
\hline 
\end{tabular}
\label{tab:models_K}  
\vspace{-2mm}
\end{table*}

Based on Section \ref{exp_viz}, later days in the temporal window generally exert stronger influence on wildfire predictions, which is intuitive given that recent conditions are typically more relevant for forecasting. However, certain variables exert significant effects much earlier. For instance, in Figure ~\ref{fig:mesogeos_lstm_avg}, \verb|tp| and \verb|lst_day| have high SHAP values as early as day 4, while in Figure ~\ref{fig:transformer_diff}b, \verb|max_temp| and \verb|min_temp| contribute meaningfully by day 3. These early signals suggest that some features encode lasting or cumulative influence, thereby extending  wildfire predictability.  Recognizing these signals strengthens early warning and long-term wildfire planning.
 % Recognizing these early signals can enhance early warning systems and strategic planning by identifying favorable wildfire conditions over longer timescales.

 % These signals indicate lasting feature effects that extend wildfire predictability, improving early warning and long-term planning.

We further evaluate how this feature importance translates into model performance (Table ~\ref{tab:models_K}). Using the Transformer architecture, we trained models on subsets of features ranked by their importance derived from SHAP. Features with high absolute SHAP values were classified as the most important, while those with values near zero were considered the least important. When trained on only ten features, the model trained on the most important subset markedly outperformed the model trained on the least important subset, with an accuracy difference of 3.75\%. Interestingly, this margin was larger than the performance gap between the full model (trained on all twenty-four features) and the model trained on the ten most important features (3.30\%). In addition to accuracy gains, prioritizing high-impact features reduced computational cost. Training time decreased by 3.86 seconds per epoch when restricting the training set to ten features, corresponding to nearly two minutes saved over a thirty-epoch schedule, while the model did not suffer significant performance degradation. Collectively, these findings underscore the value of explainability-guided feature selection: by emphasizing early and influential signals, forecasting models can achieve high accuracy at lower computational cost, ultimately supporting more timely and effective responses to extreme wildfire events.

% Based on the observations in Section \ref{exp_viz}, later days in the time window generally have a stronger influence on wildfire predictions.  
% This trend is intuitive, as more recent environmental conditions are typically more relevant and informative for forecasting forthcoming wildfire events.

% However, not all features follow this pattern uniformly. Some features begin to show a significant impact earlier in the time window. For example, in Figure \ref{fig:mesogeos_lstm_avg}, both \verb|tp| and \verb|lst_day| exhibit relatively high absolute SHAP values as early as the fourth day. Similarly, in Figure \ref{fig:transformer_diff}b, \verb|max_temp| and \verb|min_temp| already contribute meaningfully to the model's decision by the third day. These early contributions suggest that certain features have a lasting or cumulative influence that builds over time, rather than being limited to the immediate days before prediction. The results indicate forecast opportunities beyond the weather timescale, which could potentially lead to an extended predictability of wildfires. 
% Recognizing these early signals could improve early warning systems and strategy planning by identifying conditions favorable to wildfires on longer timescales.

% This goes to show that by using the right XAI tool, we can identify features that are important in predicting a wildfire event beyond the weather time scale, i.e., the seven days leading up to that event. 

\subsubsection{Results validation using LIME}
% To further validate the results we had with SHAP, we used LIME \cite{ribeiro2016should} to provide additional insights into our models' prediction. Figure ~\ref{fig:mesogeos_transformer_lime} shows that LIME explanations are aligned with SHAP values across several features. For example. \verb|smi| and \verb|tp| are shown to have positive influence on prediction of wildfires, while the early impacts of \verb|rh| and \verb|d2m| were also captured. The differences between LIME and SHAP explanations, most notably in the impacts of static features like population or land cover classes, are likely due to LIME being original designed to explain local individual predictions, rather than global trends across the entire sample space. However, the overall patterns detected by the two XAI methods are largely similar, suggesting some levels of cohesion in their approaches to interpretability.

\begin{figure*}[t]
      \centering %\{-10pt}
       \includegraphics[scale=0.18]{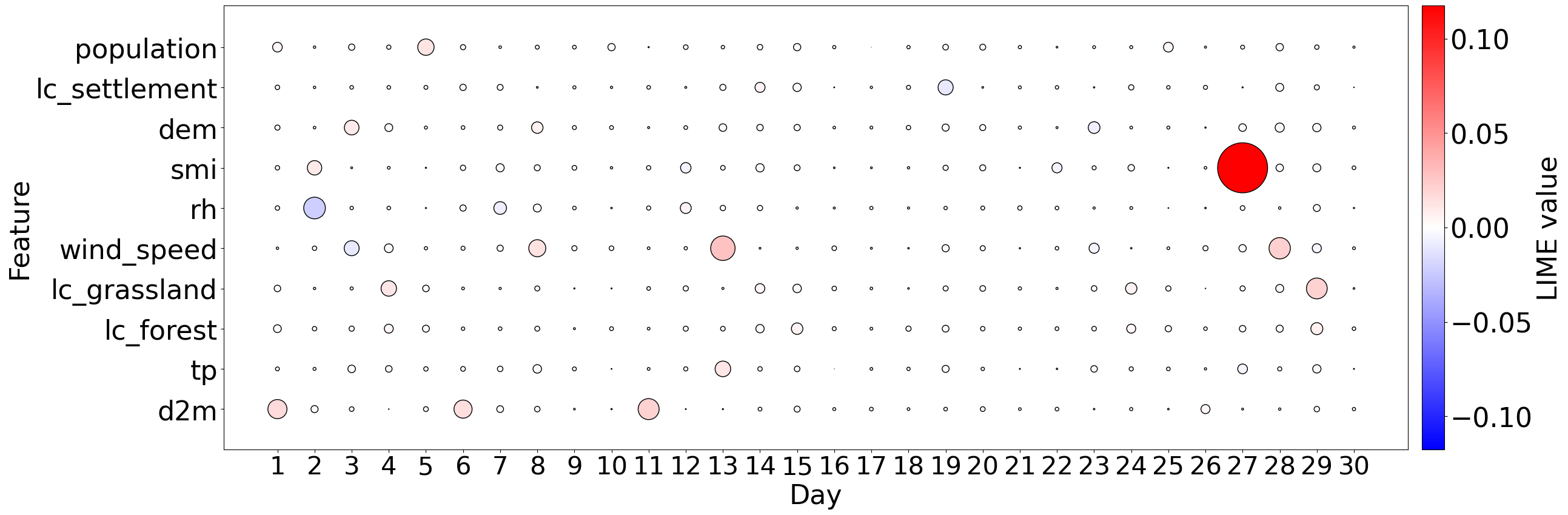} 
      \caption{Visualization for \textit{average} LIME values for the Mesogeos dataset using the Transformer model.}
      \label{fig:mesogeos_transformer_lime}
\vspace{-2mm}
\end{figure*}

To complement the SHAP-based analyses, we further employed LIME \cite{ribeiro2016should} to interrogate our models’ predictions. As shown in Figure~\ref{fig:mesogeos_transformer_lime}, the LIME explanations broadly corroborate the SHAP results across multiple features. For instance, both methods identify \verb|smi| as exerting a positive influence on wildfire prediction, while also capturing the early effects of \verb|rh| and \verb|d2m|. Notable discrepancies emerge for static features such as population and land cover classes, which is expected given that LIME was originally designed to explain local, instance-level predictions rather than global trends across the entire dataset. Nevertheless, the convergence of LIME and SHAP on key dynamic predictors underscores a degree of consistency between the two XAI approaches, reinforcing confidence in the interpretability of the models’ outputs. LIME results are also aligned with physical understanding, given that land cover types can impact the occurrence of wildfires.

\subsubsection{Implications for Mitigation and Emergency Management}
% In studying the decision-making process and performance of AI models for wildfire prediction, we are ultimately interested in determining whether current models can make physically meaningful predictions, and whether we can trust these models' decisions enough to structure our emergency response plans accordingly. The results presented in this paper point to a common theme of temperature-related variables taking on a significant role in indicating wildfire occurrence; however, there are also more subtle seasonality factors at play.

Using SHAP to interpret wildfire predictions from various AI/ML models reveals that the feature contributions to wildfire likelihood generally align with the known physical understanding of wildfire occurrence, thereby increasing confidence and trustworthiness of AI/ML predictions. The insights gained from SHAP-based explanations offer actionable opportunities for improving wildfire mitigation and emergency response strategies. Across both the Mesogeos and California Wildfires datasets, temperature-related features emerged as the most influential predictors, reinforcing the importance of monitoring thermal conditions, such as 2-meter air temperature, day- and night-time land surface temperature, and dewpoint temperature, as part of early-warning systems. %In the Mesogeos dataset, relative humidity also plays a critical role, while in California, seasonality significantly shapes the likelihood of wildfire occurrence.

By understanding how AI models respond to these factors over time, emergency management agencies can better align their decision-making processes with evolving risk conditions. For instance, deep learning models consistently identify later days in the time window as more impactful, indicating that real-time environmental monitoring in the days immediately preceding a potential fire event is crucial for accurate forecasting and rapid response. 
%However, the presence of early signals, such as elevated SHAP values for total precipitation or maximum or minimum temperatures several days before the prediction target, suggests that certain wildfire-conducive conditions can be identified well in advance. (I think this info is repeated, so I shortened it.)
In addition, the presence of early signals could serve as triggers for proactive mitigation strategies, including: 1) Pre-deployment of firefighting resources to high-risk zones, 2) Public advisories and evacuation planning in vulnerable regions, and 3) Controlled burns or vegetation management ahead of peak fire risk periods.

Moreover, seasonal patterns revealed by model explanations %(e.g., strong influence of season\_summer and season\_winter in California) 
provide guidance for long-term planning. For example, since most models agree that summer is the most fire-prone period, preparedness efforts should ramp up before summer. The different SHAP patterns in summer and winter months suggest that incorporating season-related features or temporal markers into prediction models could enhance both interpretability and effectiveness in operational settings.
%while winter conditions that may occasionally lead to wildfires require continued vigilance year-round.

%In regions like the Mediterranean, where seasonality is not explicitly encoded in datasets, model interpretability uncovers important monthly trends, such as a reversal in the contribution of dewpoint temperature from summer to winter months. This suggests that incorporating season-related features or temporal markers into prediction models could enhance both interpretability and effectiveness in operational settings.

Finally, the observed limitations of tree-based models in capturing temporal dynamics underscore the need for deploying temporal-aware architectures (e.g., LSTM, Transformer, GTN) in real-world applications where the timing of environmental signals is critical. Leveraging model explanations not only improves trust in AI predictions but also supports evidence-based policy decisions for wildfire risk mitigation and emergency response planning.

\vspace{-2mm}
\section{Conclusion}
This paper investigated how existing XAI methods can improve understanding and interpretation of wildfire prediction models. By visualizing SHAP explanations across various AI models and two geographically distinct datasets, we found a consistent emphasis on temperature-related features, seasonal-related, and precipitation, etc. as key drivers of wildfire forecasts. While this aligns with domain knowledge, the early signals revealed by SHAP values offer valuable insights for early warning and emergency planning. These insights can help authorities anticipate wildfire risk well in advance, enabling more proactive resource allocation and mitigation strategies. Overall, our findings demonstrate the potential of explainable AI to enhance wildfire prediction reliability and support more effective disaster preparedness and response.

% Our findings demonstrate the valuable role of XAI in enhancing the transparency and trustworthiness of AI models for wildfire and extreme weather prediction. This deeper insight into model decisions can guide the development of more accurate and robust future algorithms.

Following the NASA's FAIRUST principles, this work is grounded in public, verifiable data and will share code and methods to ensure the research is Findable, Accessible, Interoperable, and Reusable. Crucially, by focusing on making wildfire forecasting more Understandable through state-of-the-art XAI, we promote interpretability and reliability. Adopting FAIRUST principles advances transparency in AI for climate science while fostering open and accessible research.

\balance
\bibliographystyle{IEEEtran}
\bibliography{{IEEEfull}}

\end{document}